\documentclass[conference]{IEEEtran}
\IEEEoverridecommandlockouts

\usepackage[pdftex]{graphicx}
\usepackage{amsmath,amssymb}
\usepackage{cite}
\graphicspath{{images/}}
\usepackage{booktabs}
\usepackage{multirow}
\usepackage{float}
\usepackage{makecell}
\usepackage{adjustbox}
\usepackage{placeins}

\begin{document}

\title{Detector-Integrated Multi-Cue Association for Occlusion-Aware UAV Multi-Object Tracking%
\thanks{This research was supported by the project ``Advancing Edge-AI Enabled Airborne Systems for Maritime Intelligence and Cross-Border Collaboration'' under the University Academic Alliance in Taiwan (UAAT), funded by the Ministry of Education (MoE). Corresponding author: Trong-An Bui (trongan93@ntut.edu.tw)}%
}

\author{%
\IEEEauthorblockN{
Shu-Hung Chou\textsuperscript{1},
Trong-An Bui\textsuperscript{1}, 
Wei-Ming Chen\textsuperscript{2},
and Hsin-Piao Lin\textsuperscript{1}}
\IEEEauthorblockA{\textsuperscript{1}Institute of Aerospace and System Engineering,
National Taipei University of Technology, Taipei, Taiwan}
\IEEEauthorblockA{\textsuperscript{2}Department of Electronic Engineering,
National Taipei University of Technology, Taipei, Taiwan}
}

\maketitle

\begin{abstract}
Multi-object tracking (MOT) in unmanned aerial vehicle (UAV) imagery is challenging because vehicles are small, visually similar, and frequently occluded, which leads to identity switches (IDSW). Appearance-based trackers such as DeepSORT rely on a standalone re-identification (ReID) network, whose per-target inference limits runtime speed, while the appearance of small or occluded vehicles remains unreliable. This paper proposes a detector-integrated multi-cue tracking framework. Lightweight projection heads are attached to the neck of the YOLOv8n detector to extract target embeddings through ROIAlign, producing detections and embeddings in a single forward pass. A multi-cue association cost then combines motion, IoU, and appearance distances with weights determined through ablation, complemented by low-score rescue and appearance-memory protection. On the UAVDT benchmark, the proposed method achieves a MOTA of 21.56, a HOTA of 31.00, and an IDF1 of 37.87 at 78.11 FPS, slightly higher than those of the compared trackers. Compared with DeepSORT, it reduces IDSW by 58.7\% and runs 2.6 times faster without a standalone ReID network, although ByteTrack and OC-SORT still produce fewer IDSW.
\end{abstract}


\section{Introduction}
\label{sec:introduction}

\begin{figure}[!htb]
    \centering
    \includegraphics[width=1\linewidth]{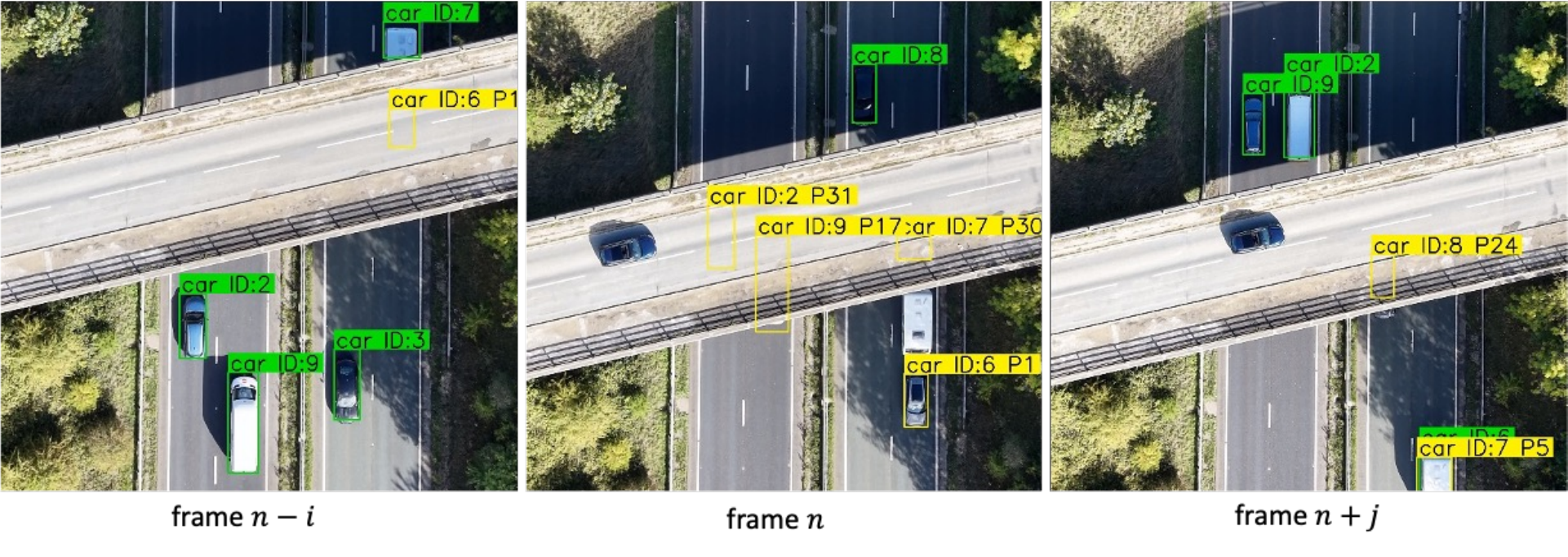}
    \caption{Illustration of identity switching caused by short-term occlusion in UAV-view object tracking. Car ID:2 and Car ID:9 are correctly tracked before entering the bridge region, but after being partially or temporarily hidden, they are incorrectly re-associated in the right frame by a conventional tracking method.}
    \label{fig:failure_case}
\end{figure}

Multi-object tracking (MOT) in unmanned aerial vehicle (UAV) imagery supports applications such as traffic monitoring, surveillance, and disaster response~\cite{du2018uavdt,YUAN10902600}. Compared with ground-view tracking, UAV-view tracking is more challenging because vehicles are small, visually similar, and densely distributed, their sizes change with flight altitude, and they are frequently occluded by bridges, trees, and buildings. When a target disappears for several frames, the tracker may assign it a new identity after reappearance, causing identity switches (IDSW)~\cite{Ho11207837}, as illustrated in Fig.~\ref{fig:failure_case}.

Existing trackers mainly follow two paradigms. Tracking-by-detection (TBD) methods first detect objects and then associate them across frames: SORT~\cite{Bewley7533003} uses Kalman prediction and IoU matching, DeepSORT~\cite{wojke8296962} adds features from a standalone re-identification (ReID) network, and ByteTrack~\cite{zhang2022bytetrack}, OC-SORT~\cite{cao2022ocsort}, StrongSORT~\cite{du2022strongsort}, and BoT-SORT~\cite{aharon2022botsort} further improve low-score association, motion modeling, and appearance representation. Joint approaches instead integrate tracking into the detector: JDE~\cite{wang2020jde} and FairMOT~\cite{zhang2020fairmot} learn detection and embeddings in a shared network, whereas joint detection and tracking (JDT) methods such as CenterTrack~\cite{zhou2020centertrack}, TrackFormer~\cite{meinhardt2021trackformer}, and MOTR~\cite{zeng2022motr} predict associations directly. For UAV-view tracking, both paradigms have limitations. Appearance-based TBD trackers require an additional ReID inference for every target, which limits runtime speed, and their association relies heavily on appearance, which is unreliable for small or occluded vehicles. JDE-style embeddings are mainly designed for pedestrians, and transformer-based JDT models are computationally expensive for real-time UAV applications.

This paper proposes a detector-integrated multi-cue tracking framework that retains the lightweight TBD association structure while obtaining embeddings in a JDE-like manner (Fig.~\ref{fig:overall_framework}). Lightweight projection heads are attached to the multi-scale neck features of the YOLOv8n detector to extract target embeddings through ROIAlign, producing detections and embeddings in a single forward pass. A multi-cue association cost then combines motion, IoU, and appearance distances with weights determined through ablation, complemented by low-score rescue and appearance-memory protection. On the UAVDT benchmark~\cite{du2018uavdt}, the proposed method achieves the best MOTA, HOTA, and IDF1 among the compared trackers; compared with DeepSORT, it reduces IDSW by 58.7\% and runs 2.6 times faster.

\section{Proposed Method}
\label{sec:methodology}

\subsection{Overview}
\label{subsec:overview}

\begin{figure*}[t]
    \centering
    \includegraphics[width=\textwidth]{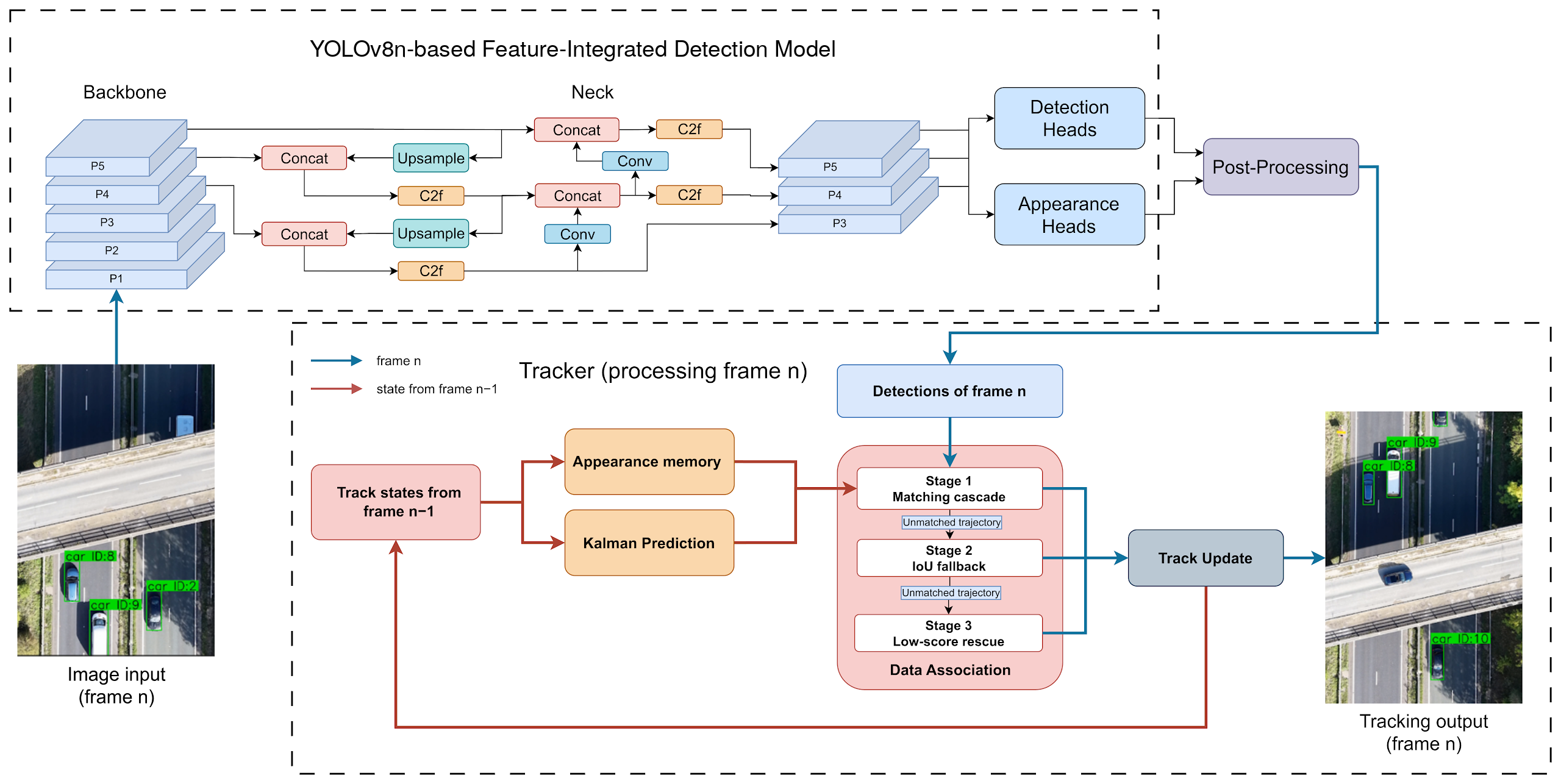}
    \caption{Overall architecture of the proposed framework. The YOLOv8n detector outputs detections and appearance features in a single forward pass. The tracker combines the appearance memory and Kalman prediction from frame $n-1$ with the detections of frame $n$ through three association stages, namely matching cascade, IoU fallback, and low-score rescue, before updating the tracks.}
    \label{fig:overall_framework}
\end{figure*}

The proposed detector-integrated multi-cue tracking framework performs online MOT on UAV imagery. It aims to reduce identity switches of small and occluded targets while lowering the computational cost of appearance extraction for real-time UAV applications. As illustrated in Fig.~\ref{fig:overall_framework}, the YOLOv8n detector outputs detections and appearance feature maps in a single forward pass, and the tracking module associates targets by combining motion, spatial, and appearance information. The framework contains two main designs. First, a detector-integrated ROI embedding attaches lightweight projection heads to the detector neck and extracts a compact appearance embedding for each target through ROIAlign, without a standalone ReID network (Section~\ref{subsec:roi_embedding}). Second, a multi-cue association cost combines the normalized motion distance with the IoU and appearance distances, reducing the reliance on appearance features and improving association stability for small and occluded targets (Section~\ref{subsec:assoc_cost}). Track prediction, association, and lifecycle management are described in Section~\ref{subsec:track_management}.

\subsection{Detector-Integrated ROI Embedding}
\label{subsec:roi_embedding}

In conventional TBD frameworks, trackers that use appearance cues, such as DeepSORT~\cite{wojke8296962}, obtain appearance features by cropping every detection and processing it with a standalone ReID network, which requires one additional network inference per target and an additional model to store and deploy. Directly reusing detector features avoids this cost; however, a detector trained only with the detection loss is encouraged to map all objects of the same class to similar representations, so pooled features of different vehicles are often too similar to distinguish identities. We therefore attach lightweight projection heads to the YOLOv8n neck and supervise them with an identity loss.

Let $P_l$ ($l\in\{3,4,5\}$) denote the neck features with strides $r_l\in\{8,16,32\}$. Each level is projected to a 64-channel map by a $1\times1$ convolution followed by batch normalization, $F_l=\phi_l(P_l)$. For a target box $b_i$, a $1\times1$ ROIAlign is applied on each level, and the results are averaged and L2-normalized:
\begin{equation}
e_i=\frac{\bar{f}_i}{\|\bar{f}_i\|_2},\qquad
\bar{f}_i=\frac{1}{3}\sum_{l=3}^{5}\operatorname{ROIAlign}\!\left(F_l,\ b_i/r_l\right),
\label{eq:roi_embedding}
\end{equation}
where $b_i/r_l$ denotes the box corners scaled to level $l$, and $e_i\in\mathbb{R}^{64}$. The same operation is applied to ground-truth boxes during training and to detected boxes after non-maximum suppression during inference.

During training, positive pairs $\mathcal{P}$ and negative pairs $\mathcal{N}$ are formed from track ID annotations, and a pairwise cosine loss is applied to the similarity $S_{ij}=e_i^{\top}e_j$:
\begin{equation}
L_{\mathrm{id}}=\frac{1}{|\mathcal{P}|}\sum_{(i,j)\in\mathcal{P}}\left(1-S_{ij}\right)
+\frac{1}{|\mathcal{N}|}\sum_{(i,j)\in\mathcal{N}}\max\left(0,\,S_{ij}-m\right),
\label{eq:id_loss}
\end{equation}
where $m$ is the similarity margin for negative pairs, set to $m=0.25$.
The positive term pulls embeddings of the same identity together, while the negative term pushes different identities below the margin. The total loss is $L_{\mathrm{total}}=L_{\mathrm{det}}+\lambda_{\mathrm{id}}L_{\mathrm{id}}$, where $L_{\mathrm{det}}$ is the original YOLOv8 detection loss, and the loss weight is $\lambda_{\mathrm{id}}=0.15$.
At inference, the detector therefore outputs each detection $d_j=(b_j,s_j,e_j)$, consisting of a box, a confidence score, and an embedding, without any additional network.

\subsection{Multi-Cue Association Cost}
\label{subsec:assoc_cost}

DeepSORT~\cite{wojke8296962} defines the association cost as $c_{ij}=\lambda d_{\mathrm{mah}}+(1-\lambda)d_{\mathrm{app}}$ and adopts $\lambda=0$ under substantial camera motion; the Mahalanobis distance then serves only as a gate, and the cost relies entirely on appearance. In UAV imagery, however, vehicles often occupy only tens of pixels, and their sizes vary considerably with flight altitude and viewing angle. When a target is small, the limited pixels inside its bounding box make the extracted appearance feature easily affected by background, blur, and box misalignment. Appearance is also unreliable under short-term occlusion: when a target is partially hidden by bridges, trees, or buildings, its box contains the occluder and the embedding is contaminated, and after reappearance the new embedding may no longer match the stored appearance memory. If the association cost relies entirely on appearance, such targets are easily mismatched or assigned new identities, as illustrated in Fig.~\ref{fig:failure_case}.

The other cues have their own weaknesses, which are complementary. For a small box, even a slight displacement causes a large drop in IoU, whereas the motion distance derived from Kalman prediction remains informative during brief visibility loss but becomes less reliable under abrupt platform motion. Since no single cue is reliable in all situations, we retain the Mahalanobis gate but, instead of relying on appearance alone, integrate motion, spatial overlap, and appearance into a unified weighted cost and seek the best combination of the three cues.

For track $T_i$ with predicted box $\hat{b}_i$ and detection $d_j$, the Mahalanobis distance is computed on the box center only. Let $\hat{\mathbf{z}}_i$ and $\mathbf{S}_i$ denote the Kalman-predicted center of $T_i$ and its innovation covariance, and $\mathbf{z}_j$ the center of $d_j$; the squared Mahalanobis distance is
\begin{equation}
d_{\mathrm{mah}}(i,j)=\left(\mathbf{z}_j-\hat{\mathbf{z}}_i\right)^{\top}\mathbf{S}_i^{-1}\left(\mathbf{z}_j-\hat{\mathbf{z}}_i\right).
\label{eq:mahalanobis}
\end{equation}
Because only the two center coordinates are used, $d_{\mathrm{mah}}$ follows a chi-square distribution with two degrees of freedom. Pairs with $d_{\mathrm{mah}}>\tau_{\mathrm{mah}}$ are rejected, and valid pairs are assigned the normalized motion distance $d_{\mathrm{mot}}=d_{\mathrm{mah}}/\tau_{\mathrm{mah}}\in[0,1]$. The spatial distance is $d_{\mathrm{iou}}=1-\operatorname{IoU}(\hat{b}_i,b_j)$, and the appearance distance is the nearest-neighbor cosine distance to the appearance memory $\mathcal{E}_i$ of the track, $d_{\mathrm{app}}=\min_{e\in\mathcal{E}_i}(1-e^{\top}e_j)$. To prevent weak appearance matches from dominating the decision, the appearance term is used only when both the spatial and appearance thresholds are satisfied:
\begin{equation}
\tilde{d}_{\mathrm{app}}=
\begin{cases}
d_{\mathrm{app}}, & d_{\mathrm{iou}}\le\tau_{\mathrm{iou}}\ \text{and}\ d_{\mathrm{app}}\le\tau_{\mathrm{app}},\\
1, & \text{otherwise}.
\end{cases}
\label{eq:gated_app}
\end{equation}
The final association cost is
\begin{equation}
C_{ij}=\lambda_{\mathrm{iou}}d_{\mathrm{iou}}+\lambda_{\mathrm{app}}\tilde{d}_{\mathrm{app}}+\lambda_{\mathrm{mot}}d_{\mathrm{mot}},
\label{eq:assoc_cost}
\end{equation}
with $\lambda_{\mathrm{iou}}+\lambda_{\mathrm{app}}+\lambda_{\mathrm{mot}}=1$. The assignment is solved by the Hungarian algorithm, and a match is accepted only if $C_{ij}\le\tau_{\mathrm{match}}=\lambda_{\mathrm{iou}}\tau_{\mathrm{iou}}+\lambda_{\mathrm{app}}\tau_{\mathrm{app}}+\lambda_{\mathrm{mot}}$, where the motion term is bounded by 1 after gating. The weighting coefficients are determined through the ablation study in Section~\ref{sec:experiments}.

Because $\tilde{d}_{\mathrm{app}}=1$ adds the full $\lambda_{\mathrm{app}}$ to the cost, a pair failing the appearance check is accepted in the cascade only if $\lambda_{\mathrm{iou}}d_{\mathrm{iou}}+\lambda_{\mathrm{mot}}d_{\mathrm{mot}}\le\tau_{\mathrm{match}}-\lambda_{\mathrm{app}}=0.237$, i.e., when motion and overlap agree closely. This is intended: a wrong match between similar neighboring vehicles is worse than a missed one, which the IoU fallback and low-score rescue can still recover. In practice the case is rare: only 0.3\% of the cascade matches on the UAVDT test sequences have $\tilde{d}_{\mathrm{app}}=1$.

\subsection{Track Management}
\label{subsec:track_management}

\textit{Track prediction:} At the beginning of each frame, an eight-dimensional constant-velocity Kalman filter predicts the state of each track, which consists of the box center, width, and height together with their velocities. The predicted box $\hat{b}_i$ and innovation covariance $\mathbf{S}_i$ are used to compute the IoU distance, the Mahalanobis gate, and the motion distance in Section~\ref{subsec:assoc_cost}.

\textit{Three-stage association:} As shown in Fig.~\ref{fig:overall_framework}, detections are first divided by confidence score into a high-confidence group ($s_j\ge\tau_{\mathrm{high}}$) and a low-confidence group ($\tau_{\mathrm{low}}\le s_j<\tau_{\mathrm{high}}$) and then pass through three association stages in sequence:
\begin{enumerate}
\item \textit{Matching cascade:} High-confidence detections are matched to confirmed tracks with the multi-cue cost in \eqref{eq:assoc_cost}. Tracks are matched level by level in ascending order of the number of frames since their last update, so that recently updated tracks take priority and tracks lost for longer, whose predictions are more uncertain, do not take their detections.
\item \textit{IoU fallback:} The remaining high-confidence detections are matched to tentative tracks and to confirmed tracks that have been unmatched for only one frame using the IoU distance alone, recovering matches missed because the appearance or motion cost is slightly too high.
\item \textit{Low-score rescue:} Following ByteTrack~\cite{zhang2022bytetrack}, low-confidence detections are not discarded. Confirmed tracks that are still unmatched and have been lost only briefly are matched to these detections using IoU alone, so that targets whose confidence temporarily drops because of occlusion or blur can keep their tracks.
\end{enumerate}

\textit{Track update and appearance-memory protection:} A track matched to a high-confidence detection corrects its Kalman state with that detection and adds the detection embedding to its appearance memory $\mathcal{E}_i$; when the memory is full, the oldest embedding is removed first. Tracks rescued in the third stage update only their position and size, and their low-confidence embeddings are not written into $\mathcal{E}_i$, which prevents features mixed with occluders from contaminating the appearance memory.

\textit{Track lifecycle:} High-confidence detections that remain unmatched initialize new tentative tracks. A tentative track is confirmed only after several consecutive successful associations and is deleted as soon as an association fails, which reduces false tracks caused by single-frame false detections. An unmatched confirmed track keeps its Kalman-predicted state and is deleted only after it has been missing for a predefined number of consecutive frames, so that a target reappearing after short-term occlusion can still be recovered. Kalman prediction, the matching cascade, IoU fallback, and the track lifecycle follow DeepSORT~\cite{wojke8296962}, low-score rescue follows ByteTrack~\cite{zhang2022bytetrack}, and appearance-memory protection is introduced in this work.

\section{Experimental Evaluation}
\label{sec:experiments}

The proposed method is trained and evaluated on UAVDT~\cite{du2018uavdt}, a UAV-view vehicle tracking dataset. The official test set contains 20 sequences (M0203, M0205, M0208, M0209, M0403, M0601, M0602, M0606, M0701, M0801, M0802, M1001, M1004, M1007, M1009, M1101, M1301, M1302, M1303, and M1401). All 20 sequences are used. The 790 ground-truth boxes of M0208 are matched by almost none of the tracker outputs (at most 2 for any tracker), indicating misaligned annotations; MOTA is therefore computed over all sequences jointly, so that this sequence does not dominate a per-sequence average. Detection is formulated as single-class vehicle detection at an input resolution of $640\times640$.

YOLOv8n is adopted as the detector because it achieves a favorable balance between accuracy and complexity among YOLO-family detectors, with 3.02M parameters, 8.2 GFLOPs, and an mAP@0.5 of 0.9916. For tracking, the projection heads and identity loss described in Section~\ref{subsec:roi_embedding} are added to the YOLOv8n detector, which is trained with UAVDT track ID annotations.

All experiments are conducted on an NVIDIA GeForce RTX 5090 GPU, with the trackers run one at a time, and the reported FPS covers the full pipeline, including preprocessing, detection, appearance extraction, association, and track update.

The fixed tracker parameters are set as follows: a track is deleted after 30 consecutive missed frames, a new track is confirmed after 3 consecutive successful associations, and each track stores up to 100 appearance embeddings. The Mahalanobis gating threshold is 5.99, the 95\% quantile of the chi-square distribution with two degrees of freedom, and the appearance-distance and IoU-distance thresholds are 0.18 and 0.65. Detections with scores of at least 0.60 are treated as high-confidence, those between 0.30 and 0.60 as low-confidence, and those below 0.30 are discarded. In the low-score rescue stage, the IoU-distance threshold is 0.45, and only confirmed tracks matched in the previous frame are considered.

Predicted boxes are matched to the ground truth with an IoU threshold of 0.5, and performance is evaluated with MOTA and IDSW~\cite{bernardin2008clearmot}, IDF1~\cite{ristani2016performance}, HOTA~\cite{luiten2021hota}, and FPS. All compared trackers use the same detector weights, test sequences, and hardware, whereas DeepSORT and BoT-SORT additionally employ standalone ReID networks.

\begin{figure}[t]
\centering
\includegraphics[width=\columnwidth]{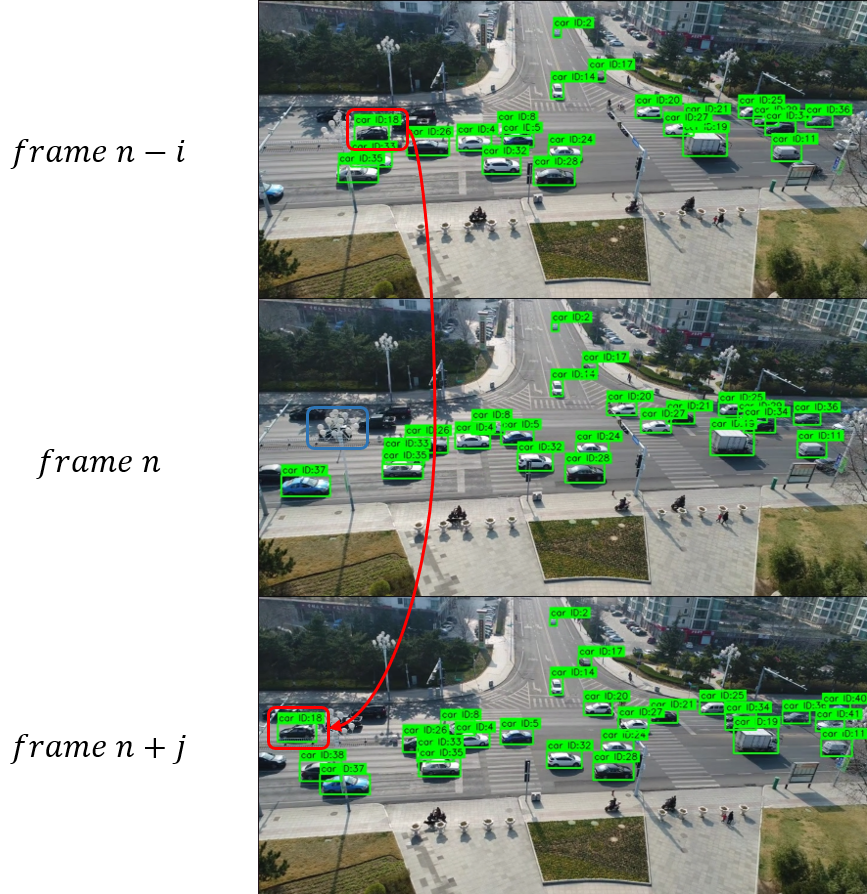}
\caption{Tracking result of the proposed method with the selected setting on the UAVDT test set. The red marking indicates a vehicle (ID:18) that is correctly re-associated after occlusion, and the blue box indicates its position while it is occluded in frame $n$.}
\label{fig:uavdt_occlusion}
\end{figure}

\begin{figure}[t]
\centering
\includegraphics[width=\columnwidth]{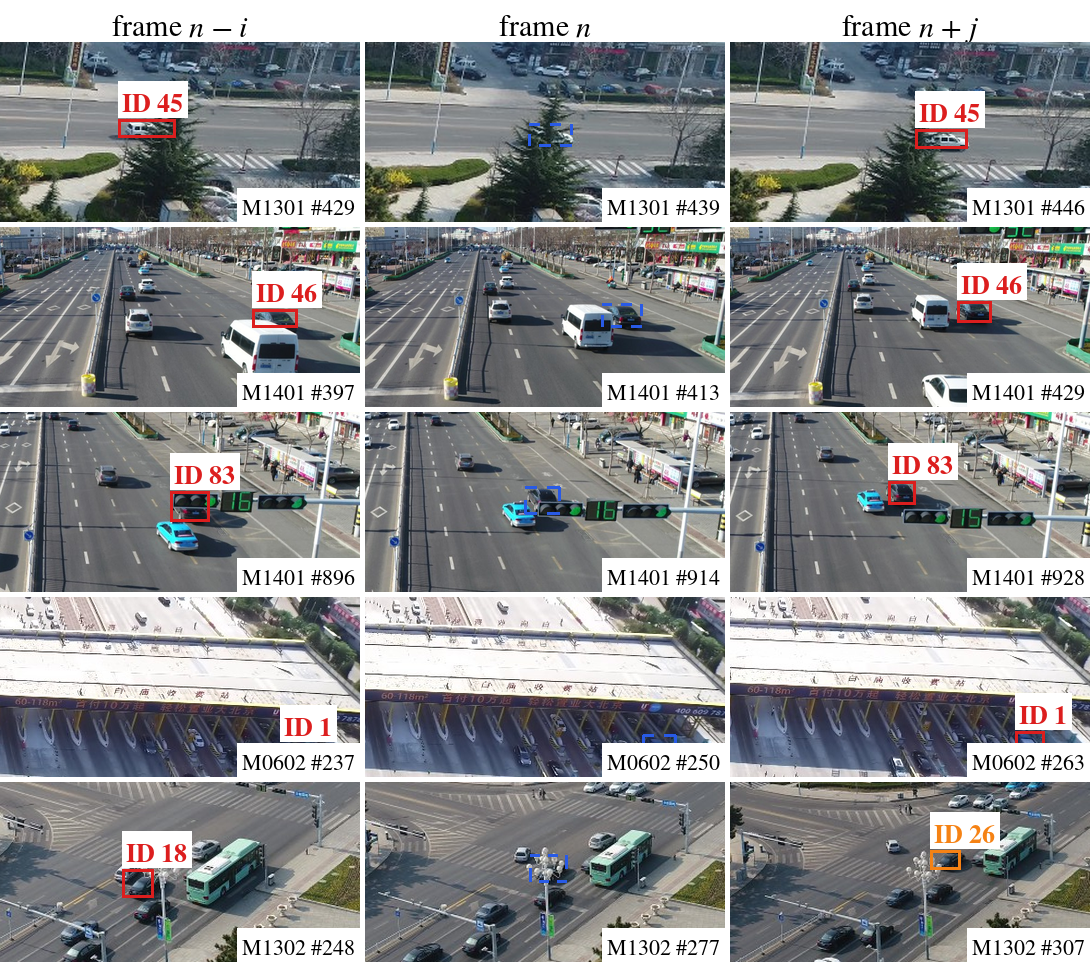}
\caption{Occlusion events from UAVDT test sequences other than M1303 (Fig.~\ref{fig:uavdt_occlusion}). Rows 1--4 keep their IDs (occluders: tree, van, traffic-light board, toll-station roof); row 5 receives a new ID after 50 occluded frames. Red: tracker output; blue dashed: ground truth while occluded; orange: ID change.}
\label{fig:occlusion_cases}
\end{figure}

\begin{figure*}[t]
\centering
\begin{minipage}[b]{0.518\textwidth}
\centering
\includegraphics[width=\linewidth]{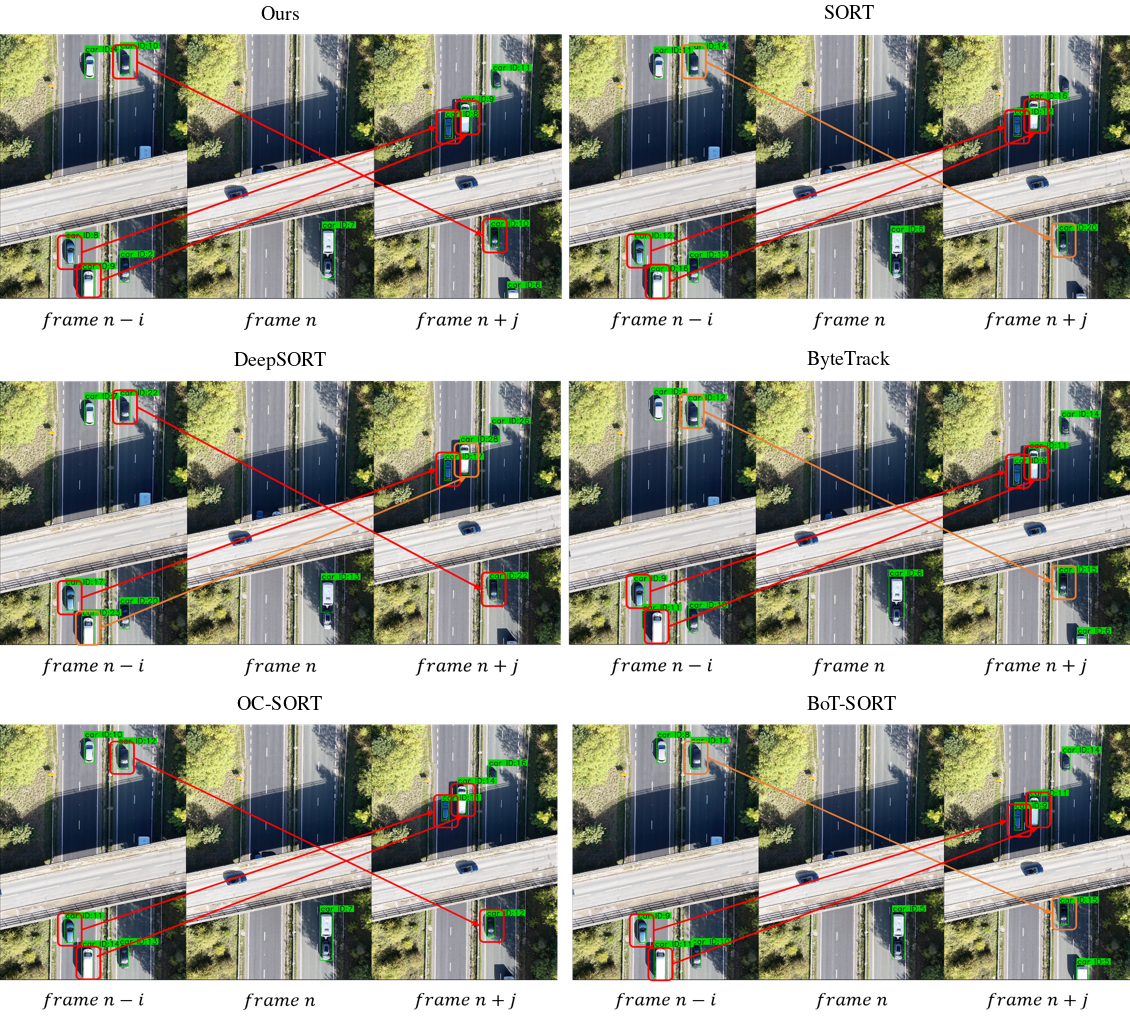}\\[-0.3mm]
\footnotesize (a) Test Video 1
\end{minipage}\hfill
\begin{minipage}[b]{0.462\textwidth}
\centering
\includegraphics[width=\linewidth]{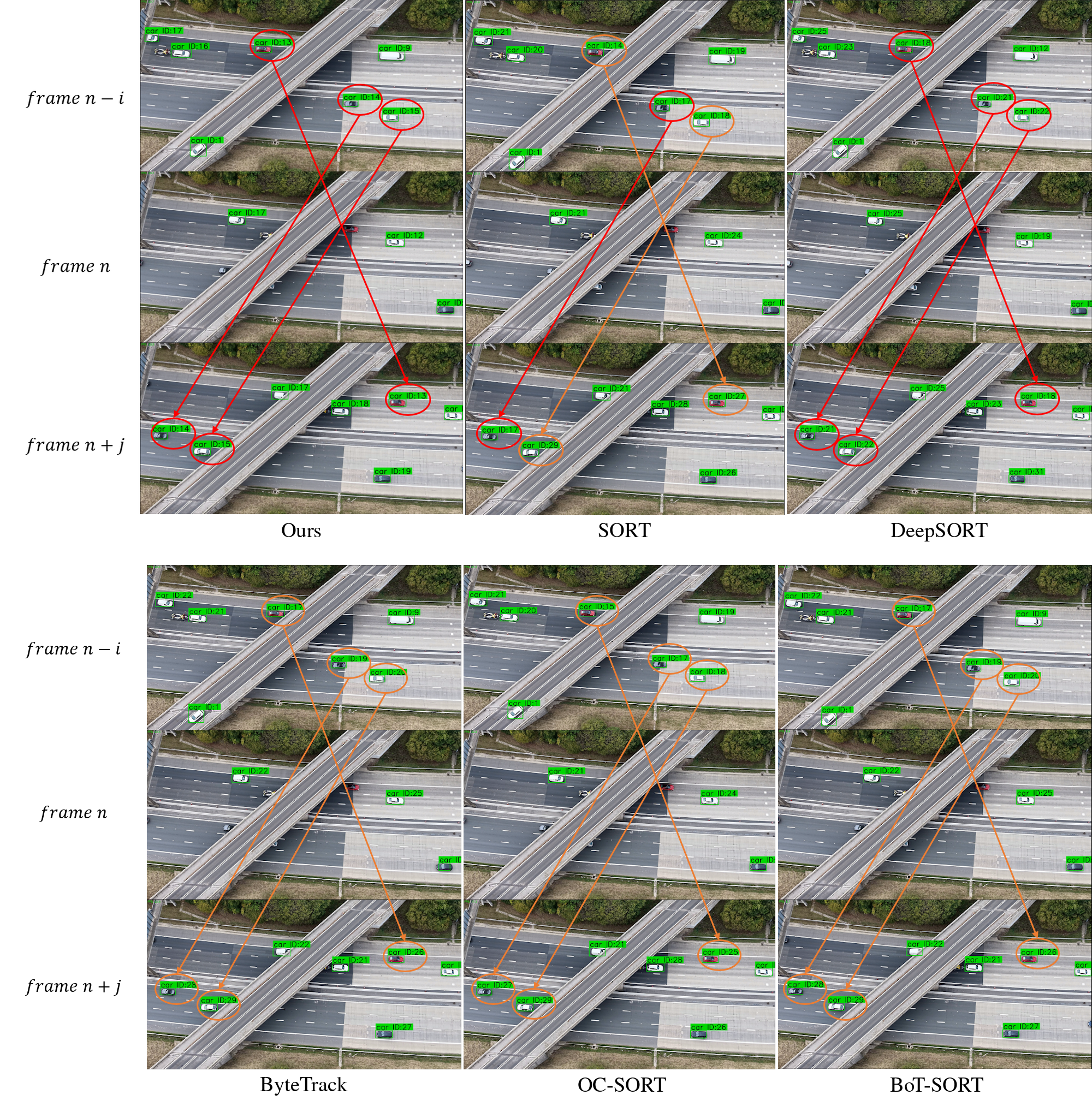}\\[-0.3mm]
\footnotesize (b) Test Video 2
\end{minipage}
\caption{Qualitative comparison of the proposed method with SORT, DeepSORT, ByteTrack, OC-SORT, and BoT-SORT on (a) Test Video 1, a multi-vehicle bridge-occlusion scene, and (b) Test Video 2, a multi-vehicle highway scene with bridge occlusion. Red markings indicate the main observed vehicle trajectories, while orange markings highlight ID changes after occlusion or reappearance.}
\label{fig:qualitative}
\end{figure*}

\begin{table}[t]
\centering
\caption{Ablation of association cues on the 20 UAVDT test sequences. Weights are listed as $\lambda_{\mathrm{iou}}/\lambda_{\mathrm{app}}/\lambda_{\mathrm{mot}}$.}
\label{tab:ablation}
\footnotesize
\setlength{\tabcolsep}{3.4pt}
\renewcommand{\arraystretch}{1.08}
\begin{tabular}{@{}llccccc@{}}
\toprule
\textbf{Cues} & \textbf{Weights} & \textbf{MOTA}$\uparrow$ & \textbf{HOTA}$\uparrow$ & \textbf{IDF1}$\uparrow$ & \textbf{IDSW}$\downarrow$ & \textbf{FPS}$\uparrow$ \\
\midrule
IoU        & 1.0/0.0/0.0 & 21.54 & 30.87 & 37.70 & 179 & 79.69 \\
App        & 0.0/1.0/0.0 & 21.51 & 30.70 & 37.35 & 224 & 77.68 \\
Mot        & 0.0/0.0/1.0 & 21.53 & 30.83 & 37.58 & 194 & 78.37 \\
\midrule
IoU + App  & 0.5/0.5/0.0 & 21.51 & 30.70 & 37.35 & 225 & 78.32 \\
IoU + Mot  & 0.5/0.0/0.5 & 21.53 & \textbf{31.00} & \textbf{37.91} & 175 & 75.41 \\
App + Mot  & 0.0/0.5/0.5 & 21.55 & 30.99 & 37.88 & 174 & 76.56 \\
\midrule
\multirow{6}{*}{\makecell[l]{IoU + App\\+ Mot}}
           & 0.1/0.2/0.7 & 21.53 & 30.99 & 37.89 & 179 & 77.42 \\
           & 0.1/0.3/0.6 & 21.54 & \textbf{31.00} & 37.90 & 174 & \textbf{80.23} \\
           & 0.1/0.4/0.5$^{\dagger}$ & \textbf{21.56} & \textbf{31.00} & 37.87 & \textbf{172} & 78.11 \\
           & 0.2/0.1/0.7 & 21.52 & 30.98 & 37.88 & 186 & 77.60 \\
           & 0.2/0.2/0.6 & 21.54 & \textbf{31.00} & \textbf{37.91} & 175 & 78.14 \\
           & 0.2/0.3/0.5 & 21.55 & 30.99 & 37.87 & 178 & 77.35 \\
\bottomrule
\multicolumn{7}{@{}l}{\scriptsize $^{\dagger}$Selected setting. Bold denotes the best value in each column.} \\[-2pt]
\multicolumn{7}{@{}l}{\scriptsize MOTA is computed over all sequences jointly.}
\end{tabular}
\end{table}

Table~\ref{tab:ablation} analyzes the three cues of the association cost in \eqref{eq:assoc_cost}. Among single cues, IoU yields the best MOTA, HOTA, and IDF1, whereas appearance alone produces the most identity switches (224), indicating that appearance is unreliable as the sole cue for small UAV-view vehicles.

Combining appearance with motion reduces IDSW to 174, the lowest among the two-cue settings, indicating that the two cues are complementary. Among the three-cue settings, $(\lambda_{\mathrm{iou}},\lambda_{\mathrm{app}},\lambda_{\mathrm{mot}})=(0.1,0.4,0.5)$ achieves the highest MOTA of 21.56, the highest HOTA of 31.00, and the lowest IDSW of 172 at 78.11 FPS. It is therefore selected as the final setting, and the result in Fig.~\ref{fig:uavdt_occlusion} is obtained with these weights on the UAVDT test set. Because UAVDT provides no separate validation split, the weights were selected on the test sequences themselves, with all other tracker parameters fixed, which may slightly favor the selected setting.

Fig.~\ref{fig:uavdt_occlusion} shows an example from UAVDT test sequence M1303. Vehicle ID:18 is occluded by a street lamp in frame $n$ (blue box) and is re-associated with the same identity in frame $n+j$ (red), showing that motion prediction and appearance memory together preserve identities through short-term occlusion. Fig.~\ref{fig:occlusion_cases} adds five occlusion events from other test sequences, selected with the per-object occlusion annotations of UAVDT: four vehicles keep their IDs after being hidden by a tree, a van, a traffic-light board, or a toll-station roof, whereas a vehicle hidden for 50 frames receives a new ID.

\begin{table}[t]
\centering
\caption{Comparison with representative trackers on the 20 UAVDT test sequences.}
\label{tab:comparison}
\footnotesize
\setlength{\tabcolsep}{4pt}
\renewcommand{\arraystretch}{1.08}
\begin{tabular}{@{}lccccc@{}}
\toprule
\textbf{Method} & \textbf{MOTA}$\uparrow$ & \textbf{HOTA}$\uparrow$ & \textbf{IDF1}$\uparrow$ & \textbf{IDSW}$\downarrow$ & \textbf{FPS}$\uparrow$ \\
\midrule
SORT~\cite{Bewley7533003}              & 20.48 & 27.41 & 31.50 & 941 & \textbf{90.16} \\
DeepSORT~\cite{wojke8296962}           & 21.15 & 30.33 & 36.80 & 416 & 30.42 \\
ByteTrack~\cite{zhang2022bytetrack}    & 20.82 & 29.75 & 36.58 & \textbf{139} & 79.86 \\
OC-SORT~\cite{cao2022ocsort}           & 20.02 & 28.64 & 34.59 & 153 & 89.65 \\
BoT-SORT~\cite{aharon2022botsort}      & 21.26 & 30.44 & 37.23 & 382 & 29.35 \\
\midrule
Ours                                   & \textbf{21.56} & \textbf{31.00} & \textbf{37.87} & 172 & 78.11 \\
\bottomrule
\end{tabular}
\end{table}

Table~\ref{tab:comparison} compares the proposed method with SORT, DeepSORT, ByteTrack, OC-SORT, and BoT-SORT. The proposed method achieves the best MOTA of 21.56, HOTA of 31.00, and IDF1 of 37.87.

Compared with DeepSORT and BoT-SORT, the proposed method reduces IDSW by 58.7\% and 55.0\%, respectively, and, because it requires no standalone ReID network, runs 2.6 and 2.7 times faster. ByteTrack and OC-SORT, which rely only on motion and overlap cues, produce fewer identity switches and run faster, but their MOTA, HOTA, and IDF1 are lower, indicating that the detector-integrated embedding improves overall identity consistency at a moderate computational cost.

Fig.~\ref{fig:qualitative} compares how well the trackers preserve identities when vehicles pass under a bridge, using two UAV videos collected from Pexels~\cite{pickup2024motorway} with three vehicles observed in each video. In Test Video 1, the proposed method and OC-SORT keep the original IDs of all three vehicles after reappearance, whereas SORT, DeepSORT, ByteTrack, and BoT-SORT each assign a new ID to one vehicle. In Test Video 2, only the proposed method and DeepSORT preserve the IDs of all three vehicles; SORT changes the IDs of two vehicles, and ByteTrack, OC-SORT, and BoT-SORT change the IDs of all three.

In these bridge-occlusion scenes, for trackers that rely only on motion and overlap cues, the reappearing detection often deviates from the Kalman-predicted position after the target has been hidden by the bridge for several frames, so it is treated as a new target. DeepSORT can recover identities with its standalone ReID network, but it runs at only 30.42 FPS. By combining motion prediction with detector-integrated appearance memory, the proposed method preserves the identities of all observed vehicles in both videos without a standalone ReID network.

\section{Conclusion and Future Work}
\label{sec:conclusion}

This paper presented a detector-integrated multi-cue tracking framework for MOT in UAV imagery. Lightweight projection heads are attached to the neck of the YOLOv8n detector, and ROIAlign is used to produce detections and appearance embeddings in a single forward pass without a standalone ReID network. On the tracking side, three-stage association is performed with a weighted cost of motion, IoU, and appearance distances, complemented by low-score rescue and appearance-memory protection, to reduce identity switches caused by small targets and short-term occlusion.

Experimental results on the 20 UAVDT test sequences show that, with the selected weights, the proposed method achieves a MOTA of 21.56, a HOTA of 31.00, and an IDF1 of 37.87 at 78.11 FPS, and all three accuracy metrics are the best among the compared trackers. The ablation study shows that appearance and motion are complementary and that combining all three cues yields the best performance. Compared with DeepSORT and BoT-SORT, the proposed method reduces IDSW by 58.7\% and 55.0\% and runs 2.6 and 2.7 times faster, respectively.

With respect to the problem addressed in this paper, the results show that the proposed method improves, to a certain extent, the association of small and scale-varying targets under occlusion. Among single cues, appearance alone produces the most identity switches, whereas combining it with motion reduces IDSW from 224 to 174. In UAVDT sequence M1303 and the two test videos, vehicles occluded by a street lamp or a bridge are recovered with their original IDs after reappearance. However, ByteTrack and OC-SORT still achieve lower overall IDSW, and new IDs are still produced under long-term occlusion or severe missed detections, indicating that occlusion-aware association still has room for improvement. In addition, the evidence for occlusion handling in this work comes from qualitative examples on UAVDT and two test videos; a quantitative occlusion-specific evaluation, further UAV benchmarks, and tests on an actual UAV platform remain to be carried out.

Future work will proceed in two directions. First, detection and tracking will be integrated into an end-to-end architecture so that association can also be learned directly by the network, with adaptive weights for the association cost and improved identity recovery after long-term occlusion. Second, the model will be made more lightweight and deployed on UAV platforms for real-time onboard tracking.

\bibliographystyle{IEEEtran}
\bibliography{bibtex/bib/IEEEabrv,bibtex/bib/IEEEmain}


\end{document}